# Will ChatGPT get you caught? Rethinking of Plagiarism Detection

Mohammad Khalil[1] and Erkan Er[2]

[1] Centre for the Science of Learning & Technology (SLATE), University of Bergen, Norway
[2] Middle East Technical University, Turkey
`mohammad.khalil@uib.no`

**Abstract.** The rise of Artificial Intelligence (AI) technology and its impact on education has been a topic of growing concern in recent years. The new generation AI systems such as chatbots have become more accessible on the Internet and stronger in terms of capabilities. The use of chatbots, particularly ChatGPT, for generating academic essays at schools and colleges has sparked fears among scholars. This study aims to explore the originality of contents produced by one of the most popular AI chatbots, ChatGPT. To this end, two popular plagiarism detection tools were used to evaluate the originality of 50 essays generated by ChatGPT on various topics. Our results manifest that ChatGPT has a great potential to generate sophisticated text outputs without being well caught by the plagiarism check software. In other words, ChatGPT can create content on many topics with high originality as if they were written by someone. These findings align with the recent concerns about students using chatbots for an easy shortcut to success with minimal or no effort. Moreover, ChatGPT was asked to verify if the essays were generated by itself, as an additional measure of plagiarism check, and it showed superior performance compared to the traditional plagiarism-detection tools. The paper discusses the need for institutions to consider appropriate measures to mitigate potential plagiarism issues and advise on the ongoing debate surrounding the impact of AI technology on education. Further implications are discussed in the paper.

## 1 Introduction

Chatbots are usually referred to as programs that can be integrated into various platforms, such as messaging apps, websites, and virtual assistants, to simulate human-like conversations. Functioned by natural language processing and machine learning techniques, chatbots try to understand and respond to user input in a conversational manner [9]. Artificial Intelligence (AI) chatbots are increasingly being used in a variety of contexts, including customer service, online shopping, entertainment, and education. These intelligent chatbots can help automate certain tasks, provide information, and improve user experience and productivity.

Recently, there have been numerous debates captivating the new AI chatbot, ChatGPT by OpenAI. ChatGPT stands for Chat Generative Pre-trained Transformer and becomes a new concept of a revolutionary AI chatbot grounded in deep learning algorithms that is designed to simulate conversation with human users over the Internet. According to recent blogs over the internet, this chatbot took the internet by storm via part of the community claiming that it will be the new Google search engine.

This powerful and easily accessible technology has recently led to concerns about plagiarism in educational settings. A recent blog article by Stephen Marche "The College Essay Is Dead" raises concerns on the usage of ChatGPT for generating massive high quality textual outputs of scholarly articles using natural language processing of chatbots [11]. Stokel-Walker [19] has highlighted that ChatGPT has great potential to provide solutions to college students on tasks such as essay writing, assignment solving, script code creation, and assessment assistance. Some counter actions have been taken for example by Australia's Queensland and Tasmania schools and New York City and Seattle school districts by prohibiting the use of ChatGPT on students' devices and networks. Many universities, colleges, and schools are evaluating similar restrictions [21]. Thus, ChatGPT can quickly become a popular choice among students to generate academic essays for homeworks, which has elevated the worries of plagiarism in academia.

Following the tempting debate on ChatGPT, this paper will bring the AI bot to further discussion from an academic perspective. In particular, we will focus on the use of ChatGPT in academic settings from the perspective of academic honesty and plagiarism. In particular, 50 different open-ended questions were prepared and asked to ChatGPT. Then the short essays generated by ChatGPT are checked for plagiarism using two popular plagiarism-detection tools, iThenticate[1] and Turnitin[2]. With some empirical evidence on the potential of ChatGPT to avoid plagiarism, this research will add new insights to the ongoing discussion on the use of AI in education. The paper is organised as follows, background in section 2, presentation of the method used in the study in section 3. Reporting of findings in section 4. Discussions followed by conclusions in section 5 and 6, respectively.

## 2 Background

### 2.1 Chatbots in Education

A chatbot is a popular AI application that simulates human-like conversations through text or voice/audio [22]. Chatbots, in response to human inquiries, provide an immediate answer using natural language as if it were the human partner in a dialogue [7]. Although the first chatbot, called Eliza, dates back to 1966 [20], the modern chatbot systems have emerged since around 2016 with a rapid increase in popularity till today [1]. Education has been one of the prominent sectors that has greatly benefited from this advancing AI technology [8]. According to a recent literature review conducted by

---

[1] https://www.ithenticate.com/ (last accessed January 2023)
[2] https://www.turnitin.com/ (last accessed January 2023)

Wollny and his colleagues [22], chatbots have been majorly used for supporting skill improvement and increasing the efficiency of education by automating some tasks, while their pedagogical role has been mostly to teach content/skill or assisting learners with some tasks. Multiple empirical studies have shown that chatbots can improve students' learning experiences and facilitate their education [9; 13].

## 2.2 ChatGPT

At the moment, ChatGPT is considered the most powerful chatbot that has ever been created [15]. Amazingly, this chatbot is capable of handling diverse tasks such as creating code snippets, performing complex mathematical operations, and creating essays, stories, and even poems. According to Rudolph, Tan and Tan [15], ChatGPT has been pre-trained on over 40 terabytes of text. In simple maths, this is close to 40 million books in a kindle format. standing for advanced Natural Language Processing (NLP) and powered by complex machine learning and reinforcement techniques, ChatGPT continues to expand and the future of this chatbot holds great promise on many aspects of our lives.

## 2.3 Cheating and proctoring

With the increase of remote assignments and tests at schools and universities, the use of online proctoring in distance education has been developing the past two decades [17]. As we witness a shift towards increased involvement of commercial entities in education, institutions forfeit control over their digital educational infrastructure [10]. This is questionable as several educational assessment models in these commercial entities might not be as trustworthy as believed, raising concerns of the reputation of academic institutions. The problem of trustworthiness is quite connected to academic cheating, which is a serious worldwide problem [2; 24]. Academic misconduct has even gone beyond distance learning entities, introducing new challenges that current teaching and learning has not experienced before. As such, a mother has discovered that her 15 years old teenager was writing her essays using a "copy robot". In China, these machines can be easily acquired with just a few clicks on the popular e-commerce website Taobao for almost one hundred United States dollars (see Fig 1).

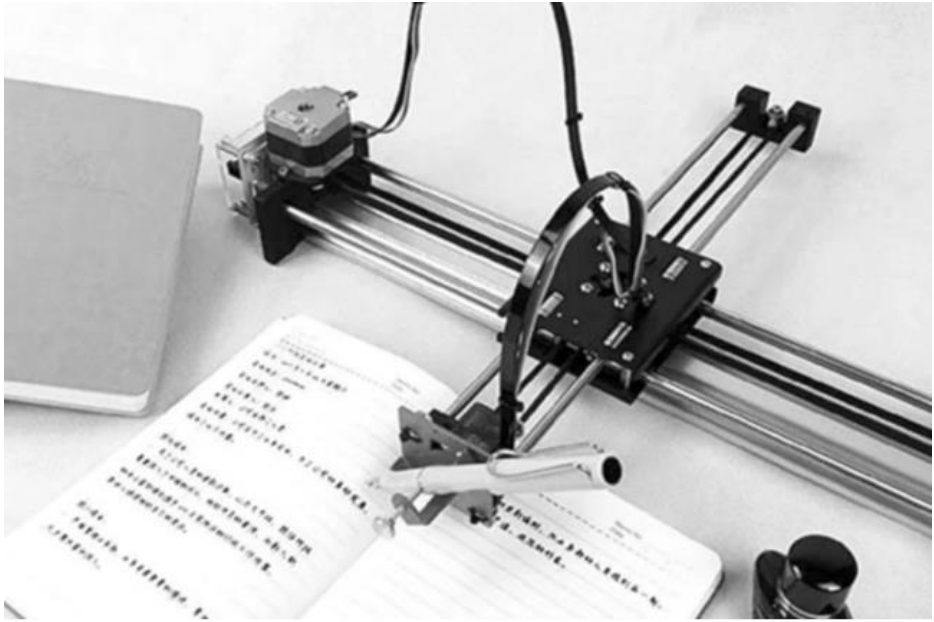

**Fig. 1.** Student got caught at school using a robot to write her daily homework [23]

### 2.4 Plagiarism check

Plagiarism involves presenting someone else's work or ideas as your own without proper attribution. Plagiarism includes not only text, but also image copying. The latter is concerned when an image or part of it is copied without a reference to the source. On the other hand, text plagiarism is what is more known in copying other people's written text. The rampant issue of online plagiarism in assignments and essays is a major challenge facing academia [4; 16].

Turnitin® and iThenticate® are two sister anti-plagiarism tools created by the same company "iParadigms LLC". Both of these anti-plagiarism tools have seen increased usage in academic institutions since 1997. According to [12; 18], iParadigms' two anti-plagiarism products have become the most popular services used to detect instances of copied work.

## 3 Methodology

This is a descriptive study that presents the results of plagiarism analysis on some content generated by AI. In particular, this study follows a quantitative analysis, where the outputs generated by a chatbot are analysed and evaluated numerically based on the originality scores produced by the anti-plagiarism tools. Below we explain in more detail the process for the data collection, plagiarism check, and further analysis.

### 3.1 Sample data and plagiarism check process

To gather a representative sample, the two authors suggested 50 different topics and instructed the ChatGPT to write "500 words essay on topic x". Each output was converted into plain text and saved into a separate file as if they were student submissions to an essay assignment on a given topic.

The collected essays were uploaded to the two plagiarism detection software. The first half was uploaded to Turnitin (n= 25) and the second half to iThenticate (n= 25). Both of the software generate a proportion of plagiarism by comparing the asked text with a massive database of internet articles, academic papers, and website pages (see Fig 2).

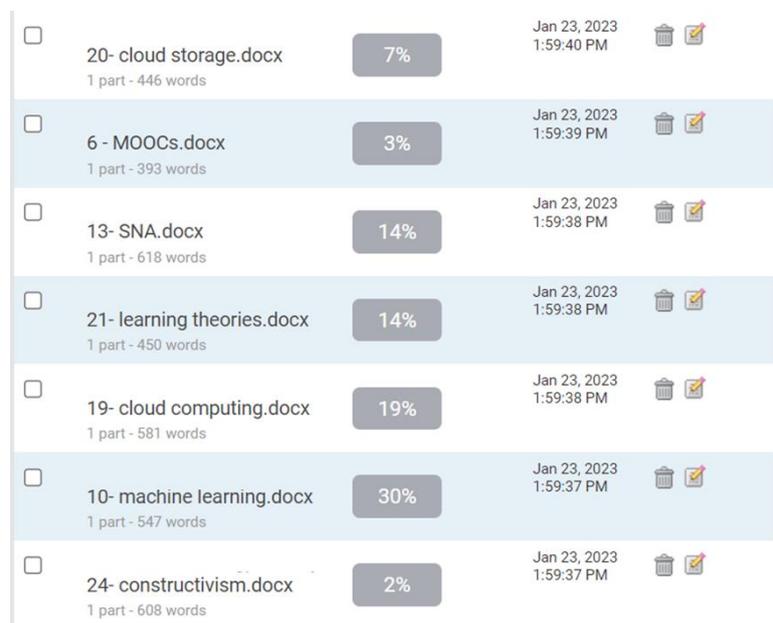

**Fig. 2.** Screenshot from iThenticate® showing the Similarity score (i.e., plagiarism proportion) of (n= 7) essays

As an additional measure in our plagiarism check, the ChatGPT bot was also used to identify any plagiarism within the essays it generated (N= 50). For this purpose, ChatGPT was given the complete list of essays, and tasked with inspecting if they were generated by itself or not. This additional step highlights the effectiveness of conventional plagiarism-detection methods to detect plagiarism within AI-generated content, compared to the AI model that generated the content.

### 3.2 Analysis

The results of the plagiarism detection were analysed to determine the originality and uniqueness of the chatbot-generated essays. The analysis is descriptive following quantitative measures of originality scores.

## 4 Findings

As each of the essays was checked against plagiarism using the two software, the similarity proportion varied between 0% and 64% of all essays. Three examples of the essays are presented in Fig 3, Fig 4, and Fig 5 showing a similarity proportion of 5%, 14%, and 64% respectively.

Robots are machines that are capable of performing tasks autonomously or semi-autonomously. They have been used for a variety of purposes throughout history, from manufacturing and assembly to search and rescue operations. Today, robots are becoming increasingly advanced and are being used in a wide range of industries, from healthcare and transportation to entertainment and education.

One of the most common uses for robots is in manufacturing and assembly. In these industries, robots are used to perform tasks that are too dangerous or difficult for humans to do. For example, robots can be used to weld, paint, and assemble products at high speeds and with a high degree of precision. They can also be used to handle hazardous materials and operate in environments that are too hot, cold, or dirty for humans to work in.

Another area where robots are increasingly being used is in healthcare. Medical robots can be used to assist surgeons in performing complex procedures, such as brain surgery or heart surgery. They can also be used to help patients with physical therapy and rehabilitation. In addition, robots are being used to perform routine tasks in hospitals, such as delivering medication and cleaning patient rooms.

Robots are also being used in transportation. Self-driving cars and drones are examples of robots that are being used to make transportation safer and more efficient. Self-driving cars can reduce the number of accidents caused by human error, and drones can be used to deliver packages and perform search and rescue operations.

In addition to these practical uses, robots are also being used in entertainment and education. For example, robots are being used to create interactive exhibits in museums and theme parks, and to help teach children about science, technology, engineering, and math (STEM).

Despite all the benefits that robots bring, there are also concerns about their impact on society. For example, there are concerns that robots will take jobs away from humans and lead to increased unemployment. There are also concerns about the safety of robots, as they may malfunction and cause harm to humans.

However, it is important to note that robots are not designed to replace human workers, but to work alongside them and make their jobs easier and safer. Moreover, with the advance in technology, many jobs that were considered to be impossible to automate are being done by robots, and this has increased productivity and efficiency.

In conclusion, robots are machines that are capable of performing a wide range of tasks, from manufacturing and assembly to healthcare and transportation. While there are concerns about their impact on society, robots have the potential to make our lives easier and safer. As technology continues to advance, we can expect to see even more uses for robots in the future.

**Fig. 3.** Example of essay-1 generated by ChatGPT on 'Robots' and plagiarism check by Turnitin shows similarity score of 5% (best seen in colour)

Learning theories are frameworks that explain how individuals acquire and process information. They provide a way for researchers and educators to understand how people learn, and to design educational programs and interventions that are effective. There are several different learning theories, each of which provides a unique perspective on the learning process.

One of the most widely-known learning theories is classical conditioning, proposed by Ivan Pavlov. Classical conditioning is a type of learning that occurs when a neutral stimulus is paired with a stimulus that naturally evokes a response. For example, the sound of a bell (neutral stimulus) can be paired with food (natural stimulus) to make a dog salivate (response) when it hears the bell. This theory suggests that learning occurs through the repeated pairing of stimuli.

Another influential learning theory is operant conditioning, proposed by B.F. Skinner. This theory states that learning occurs when behavior is reinforced or punished. Reinforcement is something that increases the likelihood that a behavior will be repeated, while punishment is something that decreases the likelihood of a behavior being repeated. For example, if a student receives praise for turning in a homework assignment on time, they are more likely to repeat that behavior in the future.

Cognitive learning theory is a perspective that emphasizes the role of mental processes in learning. This theory, proposed by Jean Piaget, suggests that learning is an active process where individuals construct their own understanding of the world by organizing and reorganizing information. This theory emphasizes the importance of understanding and addressing the cognitive processes that are involved in learning.

Social learning theory, proposed by Albert Bandura, suggests that individuals learn by observing and imitating the behavior of others. This theory focuses on the role of observational learning, which occurs when people observe and imitate the actions of others. This theory emphasizes the importance of role models, and the influence of the social context on learning.

Finally, constructivism is a learning theory that emphasizes the role of the learner in actively constructing their own understanding of the world. Constructivism suggests that people learn best when they are actively engaged in the learning process, and when they are able to connect new information to what they already know. This theory emphasizes the importance of hands-on experiences, exploration, and discovery in learning.

In conclusion, learning theories are frameworks that explain how individuals acquire and process information. There are several different learning theories, each of which provides a unique perspective on the learning process. These theories include classical conditioning, operant conditioning, cognitive learning theory, social learning theory, and constructivism. Understanding these theories can help educators and researchers to design effective educational programs and interventions that meet the needs of learners.

**Fig. 4.** Example of essay-2 generated by ChatGPT on 'Learning theories' and plagiarism check by Turnitin shows similarity score of 14% (best seen in colour)

There are many laws of physics that we observe in daily life, but some of the most fundamental and commonly observed ones are:

Newton's laws of motion: These laws describe how objects move and how forces affect their motion. The first law states that an object at rest will remain at rest, and an object in motion will remain in motion in a straight line at a constant velocity, unless acted upon by a net external force. The second law states that the acceleration of an object is directly proportional to the net force acting on the object, and inversely proportional to its mass. The third law states that for every action, there is an equal and opposite reaction.

Law of conservation of energy: This law states that energy cannot be created or destroyed, only converted from one form to another. This means that the total amount of energy in a closed system remains constant.

Law of conservation of momentum: This law states that in an isolated system, the total momentum of all objects before a collision is equal to the total momentum of all objects after the collision.

Law of thermodynamics: These laws describe how heat, temperature, and other thermal properties are related. The first law, also known as the law of energy conservation, states that energy cannot be created or destroyed, only converted from one form to another. The second law, also known as the law of entropy, states that in any energy transfer or transformation, the total entropy of the system will always increase over time.

These are some of the most fundamental laws of physics that we observe in daily life, but there are many other laws and principles that govern how the physical world works.

**Fig. 5.** Example of essay-3 generated by ChatGPT on 'Laws of physics' and plagiarism check by Turnitin shows similarity score of 64% (best seen in colour)

After the similarity scores were obtained from each plagiarism-detection software, frequency tables were created (see Table 1 and Table 2), in which the scores were grouped based on the levels of the similarity: 0-10%, 10-20%, 20-40%, and 40-100%.

According to the results obtained from the iThenticate software in Table 1, the majority of the essays (n= 17, 68%) were found to have a high originality as they were barely similar to other content (<10%). Some other essays (n= 5, 20%) had an acceptable level of similarity ranging from 10 to 20%. Only three essays were reported to have very high similarity (20-40%) with other content, and none of the articles were found to have a similarity score above 40%. The average similarity score across all essays was 8.76. From the first result set, it is clear that the essays generated by ChatGPT contained highly original content and would not face plagiarism issues if they were student submissions for an assignment.

**Table 1.** iThenticate® Plagiarism check results (n= 25 essays)

| Essay topics | Essay count (%) | Similarity score |
| --- | --- | --- |
| Cloud storage; Massive open online courses (MOOCs); constructivism; Robots; use of smartphones; Internet revolution; unsupervised machine learning; creativity; assessment in education; Natural Language Processing (NLP); Driving schools; use of chatbots in education; Technology-Enhanced Learning; self-regulated learning; online banking; leadership; spam emails; hybrid learning | 17 (68%) | <10% |
| Social Network Analysis; learning theories; cloud computing; classification in machine learning; marketing plans | 5 (20%) | 10-20% |
| Machine learning; prediction; clustering | 3 (12%) | 20-40% |
| None | 0 (0%) | >40% |
| **Total and Average** | Total (n= 25) | Average (8.76%) |

Results of the second set are presented in Table 2. At first glance, it is evident that the similarity scores were relatively higher among the second group of essays. To begin with, nearly half of the essays (n= 12) had a similarity score of less than 10%, and 6 essays exhibited an acceptable level of similarity, with scores ranging from 10% to 20%. In comparison to the first result set, where only 3 essays had similarity scores between 20-40%, a significant increase in instances of lack of originality was observed in the second set, with 6 essays displaying problematic similarity scores. Additionally, a striking case of plagiarism was identified in one of the essays, as it displayed a high similarity score of over 40% with other existing content. The average similarity score among all essays was found to be 13.72, representing an increase over the initial results set (8.76).

Table 2. Turnitin® Plagiarism check results (n= 25 essays)

| Essay topics | Essay count(%) | Similarity score |
|---|---|---|
| Kindergartens; Cultures of the Middle East and South America; Hybrid and blended teaching; Educational measurement; Difference of jobs in california and new york; Flipped vs traditional lecturing; Clustering and association rule mining; Psychologists and psychiatrists; Differential equations; PhD (Doctoral holder); Good teacher; Respiratory systems | 12 (48%) | <10% |
| Clustering algorithm; C# and Java; Data science and machine learning; Object Oriented Programing; Computer science and computer engineering; Organic chemistry | 6 (24%) | 10-20% |
| Child usage of screens; Learning Analytics and Educational Data Mining; Deep learning; Logistic regression; Global warming; Data structure | 6 (24%) | 20-40% |
| Laws of physics | 1 (4%) | >40% |
| **Total and Average** | Total (n= 25) | Average (13.72%) |

## 4.1 Reverse engineering

We also explored a reverse engineering plagiarism check on the generated essays. To do this, we asked the ChatGPT "is this text generated by a chatbot?" and then pasted the essays that had already been generated. With an accuracy of over 92%, the ChatGPT was able to detect if the written essays were generated by itself. Out of 50 essays, ChatGPT identified 46 as being plagiarised, with 4 remaining undetected as instances of plagiarism (see Fig 6).

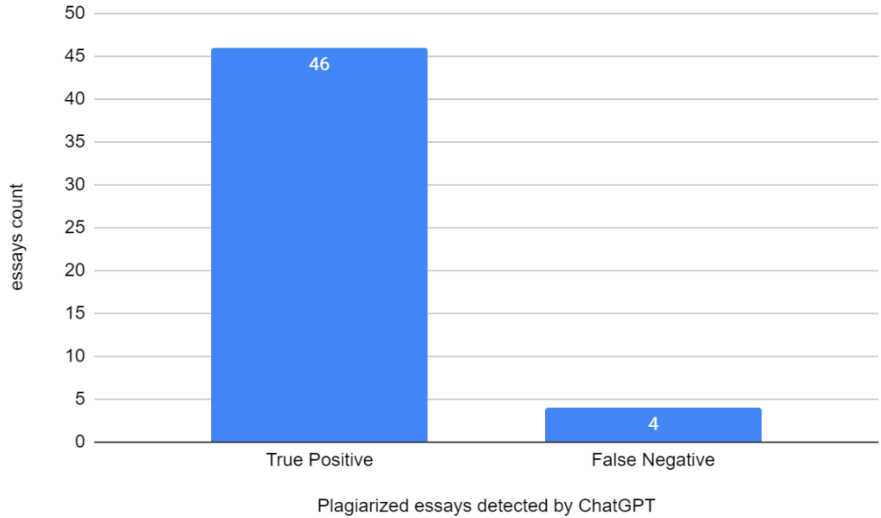

**Fig. 6.** The count of true positive (i.e., confirmed plagiarism check) and false negatives (i.e., undetected plagiarism) of the 50 essays

A response example from ChatGPT when asked about "if the text is generated by a chatbot?" is shown in the figure below (See Fig 7).

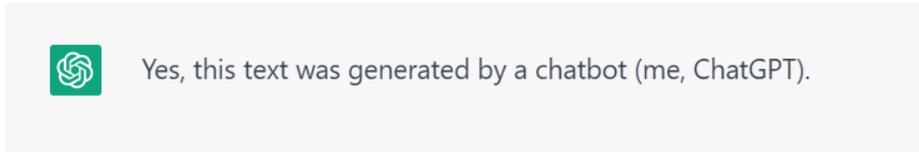

**Fig. 7.** ChatGPT answer when asked if the text is generated by a chatbot

## 5   Discussion

The findings of this study offer fresh perspectives on the ongoing debate surrounding the use of ChatGPT for academic assignments. In contrast to the 40% plagiarism rate found by Aydın and Karaarslan [3] in their evaluation of a literature review paper written by ChatGPT, our findings highlight that students may possibly use ChatGPT to complete essay-type assignments without getting caught. Of the 50 essays inspected, the plagiarism-detection software considered 40 of them with a high level of originality, as evidenced by a similarity score of 20% or less. The essays detected to plagiarism pertained to the description of various scientific topics (such as, physics laws, data mining, global warming, machine learning, etc.), which are typically considered to be factual in nature rather than interpretative. Among the essays reported with the minimum level of similarity, the topics shifted from straightforward scientific descriptions to more contentious themes that necessitated interpretation, such as cultural differences, characteristics of a good teacher, and leadership. Given the expectation for academic essays to demonstrate students' personal reflections and interpretations, it is possible that if students submit essays produced by ChatGPT, they may avoid detection by plagiarism software. Thus, this study presents compelling evidence that plagiarism with ChatGPT is already a pressing concern requiring attention. The academic community must take heed and respond proactively to address the issue at hand.

Turnitin has already raised concerns and are working on updating their plagiarism engine to detect cheating using chatbots such as ChatGPT [5; 6]. Interestingly, this study showcased an alternative solution to this issue, which involved asking ChatGPT to confirm if the given text was generated by itself or not. This approach yielded more accurate results compared to conventional plagiarism-detection tools, with 46 articles correctly predicted as generated from ChatGPT. On the other hand, plagiarism-detection tools identified only 10 essays with critical levels of similarity. These results suggest that the conventional way of detecting plagiarism has to be reconsidered and renovated in this new era of AI. Plagiarism detection may need to shift its focus from similarity check to verifying the origin of content. As evidenced by this study, possibly AI tools itself can offer a simpler yet effective solution by predicting if the text is produced by AI or not. Even the process of plagiarism-detection may need to be revised to involve a two-step approach: first, verifying the origin of the content, followed by a similarity check.

### 5.1 Study Limitation

The methodology incurs several limitations. First, the study is limited to the examination of a single chatbot technology, the ChatGPT. The results and implications may not be representative of the capabilities of all chatbot technologies, and further research may be needed to determine the generalizability of the findings. Second, the results of our study is dependent on the accuracy of the two plagiarism detection software, Turnitin and iThenticate. Third, the sample size of 50 chatbot-generated essays used in this study may not be sufficient to generalise for further implications. A larger sample size (e.g., > 1000 essays) may be necessary to increase the reliability of the results. Fourth, while we commanded the ChatGPT to generate at least 500-word essays, some of the generated texts did not adhere to this condition. Last but not least, the reverse engineering on using ChatGPT to detect plagiarism remains unverified with a similarity score such as those provided by Turnitin and iThenticate.

## 6 Conclusions

The application of large language models in education such as the OpenAI ChatGPT and Google Bard AI offer numerous possibilities to improve the educational experience for students and facilitate the tasks of teachers. Nevertheless, these chatbots may be used in an unethical way by providing students a convenient source to automatically

12produce academic essays on demand in the classroom and remote. In our study, 40 out of 50 essays composed by ChatGPT demonstrated a remarkable level of originality stirring up alarms of the reliability of plagiarism check software used by academic institutions in the face of recent advancements in chatbot technology. In response to the problem of cheating through essay generation using ChatGPT, we propose the following suggestions for the proper and effective use of ChatGPT in educational settings:

- Teachers/tutors/instructors are advised to
    - give assignments that go beyond the basics and foster active engagement and critical thinking,
    - inform students of the limitations of ChatGPT and the potential consequences of relying merely on it,
    - underline the importance of academic integrity and ethical behaviour and provide clear guidelines and expectations for students in syllabus
- Students/pupils/learners are advised to
    - take advantage of this technology as a means to improve their competencies and learning, but not as a substitute for original thinking and writing,
    - be aware of the proper and ethical use of ChatGPT in their courses and the consequences of solely relying on it for academic integrity.
- Institutions are advised to
    - get familiarised with the potentials of large language models in education [14] and open communication channels to discuss transparently with involved stakeholders, including researchers and IT support,
    - create and implement clear policies and guidelines for the use of AI tools, such as ChatGPT,
    - offer training and resources for students, faculty, and staff on academic integrity and the responsible use of AI tools in education.